\documentclass[10pt, a4paper]{article}
\usepackage{lrec}
\usepackage{multibib}
\newcites{languageresource}{Language Resources}
\usepackage{graphicx}
\usepackage{tabularx}
\usepackage{soul}
\usepackage{multirow}
\usepackage{epstopdf}
\usepackage[utf8]{inputenc}

\usepackage{hyperref}
\usepackage{xstring}

\usepackage{color}
\usepackage{authblk}
\title{CAMS: An Annotated Corpus for Causal Analysis\\ of Mental Health Issues in Social Media Posts}

\name {Muskan Garg$^{1,7}$, Chandni Saxena$^{2}$, Veena Krishnan$^{3}$, Ruchi Joshi$^{5}$,\\ \large {\textbf{Sriparna Saha$^{4}$, Vijay Mago$^{6}$, Bonnie J Dorr$^{7}$}}}

\address{$^1$Thapar Institute of Engineering \& Technology, India,\\
         $^2$The Chinese University of Hong Kong, Hong Kong SAR, \\
         $^3$University of Petroleum And Energy Studies, India
         $^4$Indian Institute of Technology, Patna, \\
         $^5$Amity University Rajasthan, India, 
         $^6$Lakehead University, Canada, 
         $^{7}$University of Florida, USA. \\
         \{muskangarg, bonniejdorr\}@ufl.edu, chandnisaxena@cuhk.edu.hk, sriparna@iitp.ac.in,\\ vkrishnan@ddn.upes.in, rjoshi@jpr.amity.edu, vmago@lakeheadu.ca \\
         }
         
\abstract{Research community has witnessed 
substantial
growth in the
detection 
of mental health issues and their associated reasons from analysis of
social media.
We introduce a new dataset for Causal Analysis of Mental health issues in Social media posts (CAMS). Our contributions for causal analysis 
are
two-fold: \textit{causal interpretation} and \textit{causal categorization}.
We
introduce an annotation schema for this task of causal analysis. 
We 
demonstrate the efficacy of our schema on two 
different datasets: (i) crawling and annotating $3155$ Reddit posts and (ii) re-annotating
the \textit{publicly available SDCNL dataset} of 1896 instances for interpretable causal analysis. We further combine 
these
into the CAMS dataset and make 
this resource publicly available 
along with associated source code:
{\url{https://github.com/drmuskangarg/CAMS}}.
We present experimental results
of models learned from CAMS dataset
and 
demonstrate that a classic Logistic Regression model outperforms the next best (CNN-LSTM) model by 4.9\% accuracy.  \\ \newline }

\begin{document}

\maketitleabstract

\section{Introduction}

With substantial growth in
digitization
of 
psychological phenomena,
automated Natural Language Processing (NLP) has 
been applied by
academic researchers and mental health practitioners to detect, classify or predict mental illness on social media. 
However, there
is a critical need for identifying 
underlying \textit{causes} of
mental illness 
in the face of dire outcomes.  For example,
a person commits suicide 
every 11.1 minutes in the US\footnote{https://suicidology.org/wp-content/uploads/2021/01/2019 datapgsv2b.pdf} and 23\% of deaths in the world are
associated with
mental disorders according to the World Health Organization.
The pandemic lockdown has 
heightened the
mental health 
crisis
in 
UK~\cite{pierce2020mental} and 
US~\cite{mcginty2020psychological}. 

\mbox{~~~}In this context, people with mental disorders 
who
decide to visit
mental health practitioners for social well-being
may
face difficulty due to
\textit{social stigma} 
or
\textit{unavailability of mental health practitioners},
leaving those 
most in need
to be neglected
by the community. As a result, sufferers of mental health conditions are 
unable to
take necessary steps for their treatment. Unfortunately, 80\% of 
those with mental health conditions
do not undergo clinical treatment and about 60\% of those who 
take their own lives previously denied
having any suicidal thoughts~\cite{sawhney2021phase}. 
Accordingly, social media platforms (e.g., \textit{Reddit, Twitter}) are
important resources for
investigating the mental health of users based on their writings.

\subsection{Motivation}

The research community 
has witnessed 
tremendous
growth in the study 
of 
mental health classification on social media 
since 2013~\cite{garg2021quantifying}. However, there is 
minimal automation for identifying 
potential causes
that underlie mental illness. 
Online users suffering from depression may
express
their
thoughts and grievances on social media unintentionally, for instance, the post $(P)$.
\begin{quote}
    $P$: \textit{I cannot deal with this breakup anymore and want to finish my life}
\end{quote}
The reason behind depression in $P$ is clearly interpreted from the word \textit{breakup},
which serves as an indicator 
of a cause related to the notion of \textit{relationship}. 
Through the application of automatic \textit{causal analysis}, underlying reasons of this type 
may
be extracted and potentially leveraged
to 
address mental health problems.  

\begin{figure}
    \centering
    \includegraphics[width=0.46\textwidth]{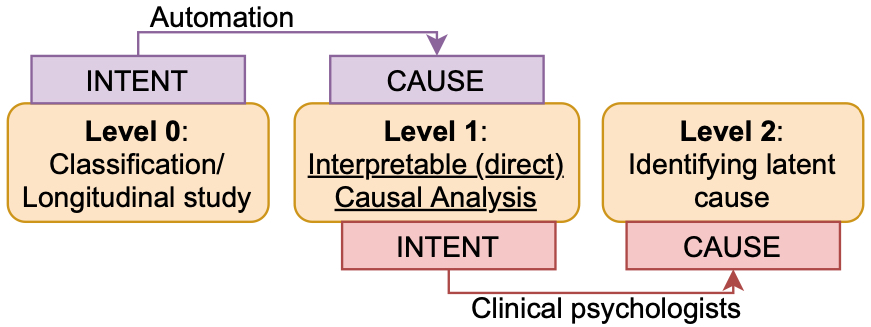}
    \caption{The intent-cause analysis of mental health on social media}
    \label{fig:overview}
\end{figure}


\begin{table*}
\centering
\scriptsize
\begin{tabular}{p{4cm}p{8cm}p{3cm}p{0.5cm}}
\hline
\textbf{Dataset} & \textbf{Details}  & \textbf{Task}  & \textbf{Avail.} \\
\hline

\textbf{CLPsych}~\cite{coppersmith2015clpsych} & Three types of annotated information: Depression-v-Control [DvC], PTSD-v-Control [PvC], and Depression-v-PTSD [DvP] & Suicide risk detection& S \\
\hline

\textbf{MDDL}~~\cite{shen2017depression} & 300 million users and 10 billion tweets in  D1: Depression D2: Non-depression, D3: Depression candidate & Depression detection  & A \\
\hline

\textbf{RSDD}~\cite{yates2017depression} & Reddit dataset of 9210 users in depression and 1,07,274 users in control group & Depression detection & ASA \\
\hline

\textbf{SMHD}~\cite{cohan2018smhd} & Reddit dataset for multi-task mental health illness & Mental health classification & ASA \\
\hline
\textbf{eRISK}~\cite{losada2018overview} & Early risk detection by  CLEF lab about problems of detecting depression, anorexia and self-harm & Depression detection & A \\

\hline
\textbf{Pirina18}~\cite{pirina2018}
&  Acquired filtered data from Reddit & Depression detection & A \\

\hline
\textbf{Ji18}~\cite{ji2018}
& Reddit: 5,326 suicide risk samples out of 20k; Twitter: 594 tweets out of 10k & Suicide risk detection & AR \\
\hline

\textbf{Aladag18}~\cite{aladaug2018} 
& 10,785 posts were randomly selected and 785 were manually annotated as suicidal v/s non-suicidal & Suicide risk detection & AR \\
\hline

\textbf{Sina Weibo}~\cite{cao2019latent}  & 3,652 (3,677) users with (without) suicide risk from Sina Weibo & Suicide risk detection& AR \\
\hline
\textbf{SRAR}~\cite{gaur2019knowledge} & Posts from 500 Redditors (anonymized) \& annotated by domain expert & Suicide risk detection& ASA \\
\hline
\textbf{Dreaddit}~\cite{turcan2019dreaddit} & 190K Reddit posts of 5 different categories & Stress detection & A \\
\hline
\textbf{GoEmotion}~\cite{demszky2020goemotions}  & Manually annotated 58k Reddit comments for 27 emotion categories & Emotion detection& A \\
\hline
\textbf{UMD-RD}\footnote{The University of Maryland Reddit Dataset}~\cite{shing2020prioritization} & 11,129 users who posted on r/SuicideWatch and 11,129 users who did not & Suicide risk detection & ASA \\
\hline
\textbf{SDCNL}~\cite{haque2021deep} & Reddit dataset of 1895 posts of depression and suicide & Suicide v/s depression classification & A \\
\hline
\textbf{CAMS} (Ours) & Interpretable causal analysis of mental illness in social media (Reddit) posts & Causal analysis& A \\
\hline
\end{tabular}
\caption{Different mental health datasets and their availability. A: Available, ASA: Available via Signed Agreement, AR: Available on Request for research work}
\label{tab:dataset_ex}
\end{table*}

\mbox{~~~}Social, financial and emotional disturbances have a huge impact on the mental health of online users. Here, we consider three levels of mental disorder analysis from \textit{automation} to \textit{latent cause} as shown in Figure~\ref{fig:overview}. We identify the \textit{intent (level 0 task)} of a user by mental illness prediction and classification of social media posts. 
We further 
automate the process of identifying and categorizing the \textit{direct cause (level 1)} 
that a user may mention in the post. 
In the future, causal analysis 
may discern 
crucial protective factors for mental health and address some important societal needs. 
Domain experts 
refer to \textit{level 1} as 
a \textit{ direct cause} mentioned by a user, 
often accompanied by 
a \textit{latent cause (Level 2)}
when they are posted on social media. In this work, we focus on automation by introducing a dataset for \textit{level 1: interpretable causal analysis}.

\subsection{Challenges and Contributions}
Mental health illness detection and analysis on social media 
presents many linguistic, technical and psychological challenges. Among many under-explored dimensions, some substantial research gaps are addressed as follows:

\begin{enumerate}
    \item Social media posts
    are first-hand user-generated data
    containing informal and noisy text. The nature of
    the
    text in a post may vary for different platforms.
    \item 
    Dataset availability 
    may be 
    limited due to 
    the sensitive nature of
    personal information. 
    \item There are many existing level 0 studies for mental health detection but no substantial study for Level 1, 
    e.g., in-depth causal analyses of disorders. 
\end{enumerate}
 \mbox{~~~}To address these challenges, we introduce the task of \textit{causal analysis}. We first introduce 
 an \textit{annotation scheme} 
 for causal analysis. The dataset annotations are carried out in two ways: (i) crawling and annotation of the Reddit dataset (ii) re-annotation of the existing SDCNL dataset~\cite{haque2021deep} for the proposed task of \textit{Causal Analysis of Mental health on Social media} (CAMS). 
 There are no existing studies for this task as observed from Table~\ref{tab:dataset_ex}. To the best of our knowledge, our work is the first 
 to address 
 \textit{causal analysis} and to provide a publicly available
 dataset for this purpose. 
 Our major contributions are:



\begin{enumerate}
    \item Definition of \textit{Interpretable Causal Analysis} and construction of an annotation schema for this new task.
    \item Annotated web-crawled Reddit dataset of 3362 instances 
    using our annotation schema.
    \item 
    Re-annotation of the existing SDCNL dataset 
    as a robustness test for our annotation schema.
    \item 
    Combination of the datasets above and introduction of our new,
    publicly available CAMS dataset.
    \item 
    Demonstration of the performance of machine learning and deep learning models using CAMS.
\end{enumerate}
Below we
discuss 
relevant background 
(Section~2) 
and introduce the annotation scheme for causal analysis 
(Section~3).
Section~4 
presents our new CAMS resource, annotation, and validation.
Annotations are verified by experts (clinical psychologist and rehabilitation counselor) and validated using statistical testing of 
Fleiss’ Kappa agreement~\cite{falotico2015fleiss}.  
We further use existing multi-class classifiers for interpretable causal analysis in 
Section~5. 
Section 6
provides
concluding remarks and future research directions.
\section{Background}
\label{sec:background}
Reddit has become one of the most widely used social media platforms.
\newcite{haque2021deep}
use two subreddits $r/depression$ and $r/suicidewatch$
to scrape the SDCNL data 
and
to validate 
a label correction methodology 
through manual annotation 
of 
this dataset
for \textit{depression} versus \textit{suicide}.
They 
also address 
ethical issues impacting
\textit{dataset availability}
and make their dataset publicly available. In this section, we discuss the \textit{evolution of mental health studies} and the \textit{historical perspective of causal analysis}.

\subsection{Evolution of Mental Health Studies}
\begin{figure}
    \centering
    \includegraphics[width=0.46\textwidth]{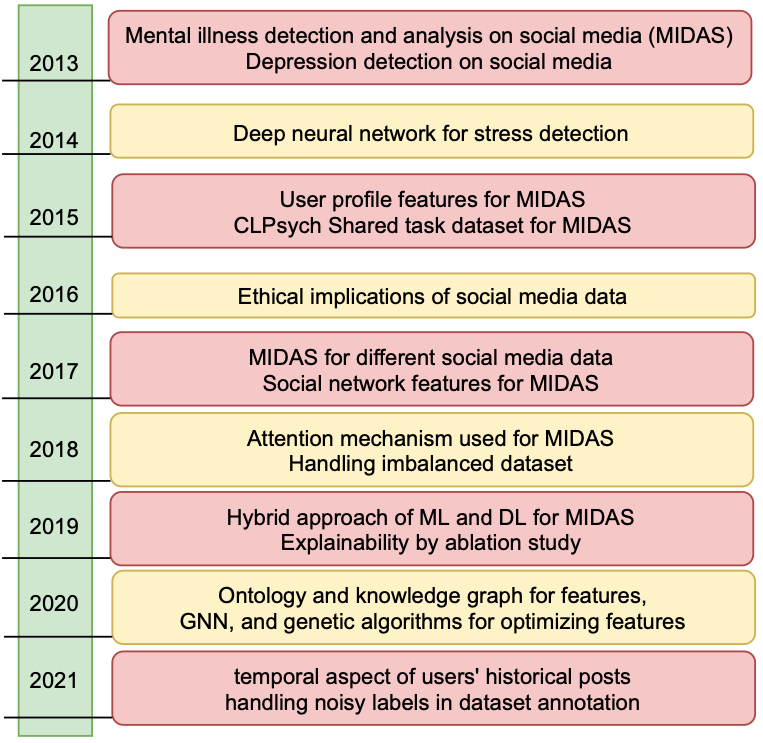}
    \caption{Evolution of studies for mental health detection on social media}
    \label{fig:timeline}
\end{figure}

Many machine learning 
researchers
have identified mental health disorders for social media posts~\cite{de2013predicting}.
\newcite{sawhney2021phase} 
examines the evolution of suicidal tendencies from historical posts of users
(longitudinal studies).
Figure~\ref{fig:timeline} highlights recent developments 
in analysis of mental health disorders from social media.

\mbox{~~~}New NLP questions have emerged from 
investigations 
into 
predicting depression~\cite{de2013predicting}
and suicidal tendencies~\cite{masuda2013suicide}.
Researchers consider users'
profiles~\cite{conway2016social} to introduce the CLPsych shared task dataset~\cite{coppersmith2015clpsych} for solving the problem of Mental Illness Detection and Analysis on Social media (MIDAS). 
MIDAS has further benefited from 
exploiting social network features~\cite{lin2017detecting}, attention mechanisms~\cite{nam2017dual}, handling imbalanced dataset~\cite{cong2018xa}, and explainability~\cite{cao2019latent}.

\mbox{~~~}Additional research directions have emerged from the
use of knowledge graphs~\cite{cao2020building}, feature optimization techniques~\cite{shah2020hybridized}, longitudinal studies~\cite{sawhney2021phase}, and handling 
noisy labels~\cite{haque2021deep}.
The 3-step theory (3ST)~\cite{klonsky2021three} of suicide supports the argument of gradual development of suicidal tendencies 
(over time) associated with a range of potential causes.

\subsection{Historical Perspective: Causal Analysis on Social Media}
Our work is relevant to 
causal analysis of human 
behavior
on social media. Recent approaches are developed to study \textit{`why online users post fake news'}~\cite{cheng2021causal}, \textit{beliefs and stances behind online influence}~\cite{Mat:22}, and \textit{causal explanation analysis on social media}~\cite{son2018causal}. The work of \newcite{son2018causal} is the closest to ours in that it detects a connection between two discourse arguments to extract a causal relation
based on annotated Facebook data. However, the dataset is limited and is not 
publicly available; thus, no recent developments are observed. To
address this issue,
we annotate the Reddit dataset (the nature of Reddit data is different from that of Facebook) and further categorize causal explanations.

\section{Annotation Scheme}
\label{sec:annotation}
\subsection{Inferences from Literature}

Potential reasons behind
mental illness may be detected in posts that refer to insomnia, weight gain, or other indicators of worthlessness or excessive or inappropriate guilt. Underlying reasons may include:  
\textit{bias or abuse}~\cite{radell2021impact},
loss of \textit{jobs or career}~\cite{mandal2011job},
physical/emotional illness leading to, or induced by, \textit{medication} use~\cite{smith2015depression,tran2019depression},
\textit{relationship} dysfunction, e.g., marital issues~\cite{beach2002marital}, and
\textit{alienation}~\cite{edition2013diagnostic}. 
This list is not exhaustive, but it is a starting point for our study, giving rise to five categories of reasons (plus `no reason') for our automatic causal analysis: \textit{no reason}, \textit{bias or abuse}, \textit{jobs and careers}, \textit{medication},\footnote{We recognize `medication' as both an \textit{indicator} of physical/emotional illness (e.g., an intent to alleviate illness) and a potential \textit{cause} of illness (e.g., medication-induced depression).} \textit{relationship}, and \textit{alienation}.

\subsection{Annotation Task}

Table~\ref{tab:examples} presents examples of data annotation involving the labeling of \textit{direct causes} of mental health disorders in social media posts.
There are two types of annotations: \textit{cause category} and \textit{Inference}. The \textit{Inference} column contains textual data which represents the actual \textit{reason} behind mental disorders. This \textit{inferred reason} is further classified as one of the six different causal categories.
\begin{figure}
    \centering
    \includegraphics[width=0.3\textwidth]{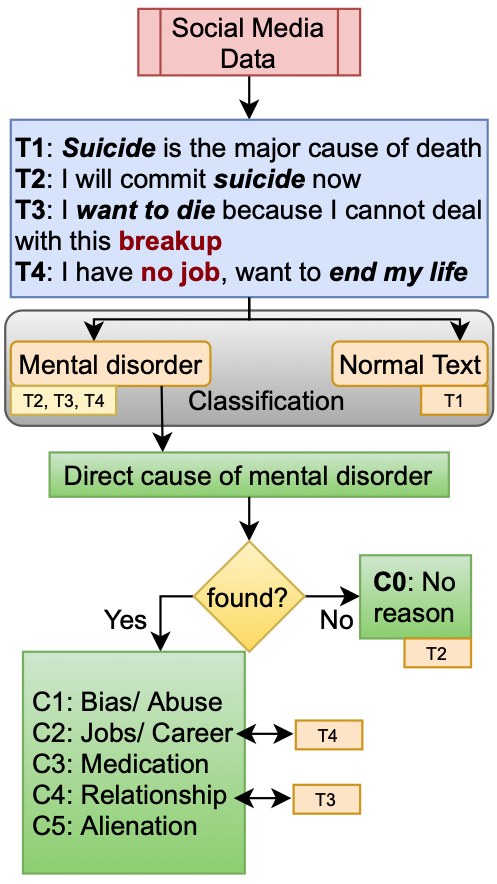}
    \caption{Architecture of the causal analysis for mental health in social media posts}
    \label{fig:1}
\end{figure}

\begin{table*}[]
    \centering
    \scriptsize
    \begin{tabular}{p{0.60\textwidth}p{0.15\textwidth}p{0.15\textwidth}}
        \hline
        \textbf{Text} & \textbf{Cause} & \textbf{Inference}\\ 
        \hline
         That's all I can really say. Nothing is worth the effort... I don't think I am capable of taking steps to improve my life, because I just don't even fucking care. Whatever I'm doing, wherever I go in my life, I'll find a way to be miserable. Why try... I just... ugh... & Alienation & Nothing is worth the effort\\ 
         \hline
          
          Does anyone feel like the only person that could understand your depression would be someone else that was depressed? But also feel like if they were to date someone who was depressed they couldn't handle it because it might suck you into a place that you don't want to be in again. & No reason & - \\
          \hline
          God help me.... I know I should go to the hospital.  I know I have to keep fighting....if only to prove to my children, cursed with these genetic tendencies of mine, that life is worth living.  I made my son promise at Christmas to get help, and he did and he is thriving.  My life long battle is starting to wear on this old soul of mine.  It feels like the same pattern over and over, no matter how many variables I change.  I am a very hard person to love.  My scars and cynicism are just a little too hard for anyone who tries to stay around too long.  & Medication & go to the hospital, scars and cynicism, genetic tendencies \\
          \hline
          I hate my job .. I cant stand living with my dad.. Im afraid to apply to any developer jobs or show my skills off to employers..I dont even have a car rn... I just feel like a failure..Im really lonely.. I feel like everything is getting worse and worse.. maybe I should start taking anti depressant medication.. & Jobs and Careers & hate my job, feel like failure, really lonely\\
          \hline
          it doesn't make any sense. can figure out what i may have done to trigger it, but 5 of my closest friends from high school have stopped responding to my calls or texts. i thought it was just a phone issue at first, but it is too unlikely of just coincidence. & Relationships & 5 of my closest friends, stopped responding \\
          \hline
          I was with a group of friends last night, and another friend started talking to another friend about how many girls secretly liked him and stuff. It was crazy, because no one has ever talked to me about things like that. Then, on the way to the pub, a group of girls basically called me unattractive. Funny how girls are never shy about calling me ugly, but they're apparently too shy to ``approach me".& Bias or Abuse & girls call me unattractive\\ 
          \hline
    \end{tabular}
    \caption{Sample of the CAMS dataset for causal analysis}
    \label{tab:examples}
\end{table*}

\begin{table}[]
    \centering
    
    \begin{tabular}{p{0.1\textwidth}p{0.3\textwidth}}
    \hline
    \textbf{Range} & \textbf{Interpretation} \\
    \hline
         $<0$: & Less than chance agreement\\
        0.01–0.20: & Slight agreement\\
    0.21– 0.40: & Fair agreement\\
    0.41–0.60: & Moderate agreement\\
    0.61–0.80: & Substantial agreement\\
    0.81–0.99: & Almost perfect agreement\\
    \hline
    \end{tabular}
    \caption{Interpretation of resulting values of Fliess' Kappa agreement study
    .}
    
    \label{tab:kappa_sig}
    
\end{table}

\subsection{Problem Formulation}
\label{problem}
The architecture for our automatic causal analysis is shown in Figure~\ref{fig:1}. 
Social media text is provided to prediction/classification algorithms that filter out \textit{non-mental disorder} from posts. The remaining \textit{mental disorder} posts are then analyzed to detect reasons behind users' depression or suicidal tendencies. Finally, the reasons are classified into 5 causal categories and one `no reason' category.

\mbox{~~~}More formally, we
introduce the problem of Causal Analysis of Mental health on Social media (CAMS) as a multi-class classification problem. We extract a set of social media posts as $p={p_1, p_2, p_3, ..., p_n}$ for $n$ number of posts. We \textit{interpret the cause} for every $i^{th}$ post $p_i$ as $C_{p_i}$ and \textit{classify} it into one of the predetermined categories $y={y_0, y_1, y_2, y_3, y_4, y_5}$ where $y_0$: `no reason', $y_1$: `bias or abuse', $y_2$: `Jobs and careers', $y_3$: `Medication', $y_4$: `Relationship', and $y_5$: `Alienation' as $y_{p_i}$.

\subsection{Guidelines for Annotation}
\label{sec:guidelines}

Our
professional guidelines support
annotation of
the post $p_i$ with \textit{causal inference} $C_{p_i}$ and 
class $y_{p_i}$. The guidelines are developed 
through
collaborative efforts 
with a clinical psychologist and a rehabilitation counselor. 
Student annotators label
posts with their \textit{causal inference} and \textit{causal class}. The latter are annotated as follows:
\begin{enumerate}
    \item \textbf{No reason}: When there is no reason that identifies the cause of mental disorder in the post. 
    \begin{quote}
        $C_{y_0}$: [``I just want to die", ``Want to end my life".]
    \end{quote}
    \item \textbf{Bias or abuse}: A strong inclination of the mind or a preconceived opinion about something or someone. To avoid someone intentionally, or to prevent someone from taking part in the social activities of a group because they dislike the person or disapprove of 
    their
    activities. It includes body shaming, physical, sexual, or emotional abuse.
    
    \begin{quote}
         $C_{y_1}$: [``No one speak to me because I am fat and ugly.", ``It has been 5 years now when that horrible incident of ragging shattered down all my confidence".]
    \end{quote}
    
    \item \textbf{Jobs and careers}: Financial loss can have catastrophic effects on mental illness, relationships and even physical health. Poor, meaningless and unmanageable education, unemployment, un-affordable home loans, poor financial advice, and losing a job are some of the major concerns. It includes gossiping and/or social cliques, aggressive bullying 
    behavior, poor communication and unclear expectations, dictatorial management techniques that don’t embrace employee feedback. The educational problems like picking up courses under some external pressure and poor grades are also part of this category.
    
    \begin{quote}
        $C_{y_2}$: [``cant afford food or home anymore", ``I do not want to read literature but my parents forced me to do so. Not happy with my grades"]
    \end{quote}
    
    \item \textbf{Medication}: The general drugs and other antiviral drugs can increase the risk of depression. The habit of using substances and alcohols can aggravate the problem of mental disorders. Moreover, medical problems like tumors, cancer, and other prolonged diseases can boost the presence of mental depression. 
    
    \begin{quote}
        $C_{y_3}$: [``I am chain smoker, want to quit, but I cant. My life is mess", ``tried hard to leave drugs but this dire craving is killing me.."]
    \end{quote}
    
    \item \textbf{Relationships}: When two people or a group of people fight, it may lead to a relationship or friendship drifting apart, for example, regular fights, breakups, divorce, mistrust, jealousy, betrayal, difference in opinion, inconsistency, conflicts, bad company, non-commitment, priority, envy. Problems like bad parenting and childhood trauma are also part of this category. 
    
    \begin{quote}
        $C_{y_4}$: [``Cannot deal with this breakup anymore", ``He dumped me and its killing me"]
    \end{quote}
    
    \item \textbf{Alienation}: Alienation is the feeling of life being worthless
    even after doing everything. 
    There may be indicators of
    meaninglessness, loneliness, tired of daily routines, powerlessness, normlessness, isolation, and cultural estrangement. 
    \begin{quote}
        $C_{y_5}$: [``I don't know why am I living, everything seems to be meaningless"]
    \end{quote}
    
\end{enumerate}
The student annotators were trained by experts (clinical psychologist and rehabilitation counselor) to pick those words and/ or phrases through which they have identified the $y_{p_i}$ of the post $p_i$, and rank them. The student annotators followed these guidelines thoroughly.  

\subsection{Annotation Perplexity}
\label{sec:annotate}
The judgement of reasons behind online users' intentions
is a complex task for human annotators,
generally due to mentions of \textit{multiple reasons} 
or the presence of \textit{ambiguity in human interpretations}. 
Causal analysis can be
viewed as
a multivariate problem, resulting in
multiple labels. The annotation scheme is not sophisticated enough to capture all the aspects of this phenomenon. 
We thus propose 
perplexity guidelines 
to simplify the task and facilitate future annotations. 
Our mental health therapists and social NLP practitioners 
have constructed 
perplexity guidelines to handle the trade-off between 
task complexity 
and simplicity of the annotation scheme. 
The \textit{perplexity} guidelines are:

\begin{enumerate}
    \item \textbf{Multiple reasons in the post}: There are some posts with multiple reasons 
    for conveyed feelings. 
    To resolve this, annotators must find
    a \textit{root cause} among the \textit{direct causes} mentioned by the user.
    \begin{quote}
        Example 1: I was of 11 years since when i realized and facing constant ignorance of my parents. 8 yrs later i lost my first girlfriend and alcoholic since then. My beer belly and obesity has made people biased towards me. I have lost everything and want to end up now.
    \end{quote}
    In Example 1, the root cause of the mental disorder 
    is \textit{ negligence of parents}. Thus, 
    this post is assigned 
    the \textit{Relationship} category. That is, 
    we handle multiple causes by prioritizing the root cause or the most emphasized reason by the user. This \textit{perplexity guideline} 
    reduces
    annotation ambiguity
    and helps develop better models for automation of this task in the near future.  
    \item \textbf{Ambiguity in human interpretations}: The subjective nature of causal analysis makes this task even more complicated for human annotators. The six different causes are not atomic in nature. 
    The human interpretation 
    of the same post and the same inference may vary, even among experts.
    \begin{quote}
        Example 2: I wish I could stay alone somewhere and cry my self to sleep. I wish i won't wake up.
    \end{quote}
    Example 2 contains some important words like \textit{alone} and \textit{cry} which convey the category as \textit{Alienation}. However, 
    two out of three annotators considered this to be the
    \textit{No reason} category. As this is the subjective decision of every human annotator, we leave it at their discretion. 
    However, the final category assigned by the human expert (following human annotation) is based on ``majority rules,'' in this case, `no reason.'
    \item \textbf{Subject of intent in the post}: 
    Many posts 
    refer to the depression of 
    loved ones and other acquaintances. Given that the
    goal
    of this task is to identify the cause behind mental depression of online users, experts agree that such posts are candidates for causal analysis.
    \begin{quote}
        Example 3: I love to do prepare meals for my cousin because I think he is suffering from depression due to his car accident last month.
    \end{quote}
    In Example 3, the user is talking about
    their cousin who is purportedly 
    suffering from depression. This text 
    is passed through classification to detect \textit{depression}; however, the third person usage precludes detection of a reason behind this condition.
    Although the user presents their own perspective on the reason for their cousin's condition (car accident), the input must include self-reported evidence for the reason. Thus, this example is 
    annotated
    as \textit{No reason}.
\end{enumerate}

The professional training and guidelines are supported by perplexity guidelines. We have further deployed student annotators to 
label
the dataset after they annotate the first $25$ posts under the supervision of experts. 




\section{CAMS Dataset}
\label{sec:CAMS}
We introduce a new language resource for CAMS
and elucidate the process of data collection~(4.1) and data annotations~(4.2). We further discuss the challenges and discuss 
future research directions (4.3).
We make our dataset and the source code publicly available for future use.
\subsection{Overview of Data Collection}
\label{sec:method}
We 
demonstrate
the efficacy of our annotated scheme 
as follows:
\begin{enumerate}
    \item We 
    collect $3362$ Reddit posts which are available with subreddit r/depression using Python Reddit API Wrapper~(PRAW). 
    Experts remove empty and irrelevant instances from the crawled dataset resulting in $3155$ samples.
    \item We leverage
    the existing SDCNL dataset comprised of $1,896$ posts: $1517$ training samples and $378$ testing samples, assumed
    as the cleaned dataset. 
    \item
    We combine these two 
    corpora, introducing them as the CAMS dataset, which is further annotated by our trained student annotators.
    \item We 
consult
mental health practitioners, a clinical psychologist and a rehabilitation 
counselor,
to verify the 
combined
dataset. 
 
\end{enumerate}



\begin{table*}[t]
    \centering
    \begin{tabular}{l|ccc|ccc|ccc}
        \hline
         \textbf{Class} & \multicolumn{3}{c}{\textbf{Crawled corpus}} &
            \multicolumn{3}{c}{\textbf{SDCNL Training data}} &
             \multicolumn{3}{c}{\textbf{SDCNL Test data}}               \\
        
          & \textbf{Min} & \textbf{Max} & \textbf{Avg} & \textbf{Min} & \textbf{Max} & \textbf{Avg} & \textbf{Min} & \textbf{Max} & \textbf{Avg} \\
          \hline
         \textit{No reason} & 1 & 508 & 59.78 & 1 & 1785 & 68.58 & 1 & 1562 & 84.85   \\
         \textit{Bias or Abuse} & 6 & 2109 & 347.48 & 5 & 4378 & 227.24 & 6 & 578 & 149.80  \\
         \textit{Jobs and career} & 13 & 2258 & 228.28 & 17 & 2771 & 255.70 & 20 & 1481 & 206.95   \\
         \textit{Medication} & 5 & 1552 & 213.83 & 3 & 3127 & 205.86 & 11 & 1124 & 165.60  \\
         \textit{Relationship} & 2 & 3877 & 229.35 & 14 & 2739 & 240.08 & 9 & 756 & 202.56 \\
         \textit{Alienation}  & 3 & 1592 & 153.86 & 1 & 899 & 147.01 & 12 & 683 & 145.67 \\
         \hline
    \end{tabular}
    \caption{Word 
    length variation in posts 
    across causal classes
    for each 
    dataset.
    }
    \label{tab4}
\end{table*}
\subsection{Dataset Annotations}
\label{sec:secAP}

After verification of the dataset by experts, 
three duly trained student annotators
manually annotate the data in the format: $<$text, cause, inference$>$ as shown in Table~\ref{tab:examples}. In this section, we discuss the annotation process, verification by experts, and validation using statistical tests. 

\mbox{~~~}Annotation is carried out manually by
annotators who are proficient in the language. 
They work independently for each post and follow the given guidelines. Each annotator 
takes
one hour to annotate about $15-25$ Reddit posts and $180$ and $90$ hours to annotate the \textit{crawled} and \textit{existing} dataset, respectively. The annotations are obtained as three separate files.

\mbox{~~~}Several challenges have emerged during the annotation process, e.g.,
a countable 
($<10$) set of non-English posts.
For such cases, 
augmented guidelines instruct the annotator
to mark 
the post as \textit{No reason}. 
We recommend the removal of non-English posts as the CAMS dataset is proposed for English 
only. The annotated files are verified by a clinical psychologist and a rehabilitation 
counselor. This verification is performed over the annotations given by our trained annotators without bringing this to their knowledge and experts have given the final annotations. 

\mbox{~~~}We further validate three annotated files using Fliess' Kappa inter-observer agreement study. The agreement study for the crawled dataset is found to be $64.23\%$. We also study the inter-agreement annotations for the existing dataset as $73.42\%$ and $60.23\%$ for testing and training data of SDCNL, respectively. The trained annotators agree with $61.28\%$ agreement among the annotators for the CAMS dataset. The resulting values are interpreted as per Table~\ref{tab:kappa_sig}. Despite the increased subjectivity of the task, the student annotators \textit{substantially agree} with their judgements.   

\subsection{Discussion}
Existing work 
on 
causal analysis is associated with finding discourse relations among words and identifying the segments 
that 
represent the reason behind the intended information. We 
extend
this work to find the category of the cause using these interpreted segments.
Since we are extending the causal interpretations to automatic categorization, our work is not directly comparable 
to 
any of the existing works. 
However, we glean insights into the
characteristics of the
CAMS dataset 
through further analysis.
This section 
examines the 
word 
length of 
posts in
the
dataset~(4.3.1) 
and varying number of instances in 
each
class~(4.3.2). 
Additionally, we discuss the social nature of the dataset~(4.3.3).

\subsubsection{Length of the Posts 
}
\label{sec:len}
The length of posts varies from a few characters to thousands of words. One of the major challenges for automation 
is the construction of
a multi-class classifier 
that
is suitable for 
posts of varying 
word
lengths. One of the possible solutions 
to this challenge is to extract the inference from the post and classify it using the inference text. We choose to explore this in the near future. 
Table~\ref{tab4}, 
shows
that, although there is consistency in the average number of words among all the classes, there is a huge variation in the word counts across posts.
This shows that the data is unstructured and to handle this kind of text, we need handcrafted features or other semantic measures for causal analysis.

\subsubsection{Imbalanced Dataset}
\label{sec:imbalanced}
Table~\ref{tab:cause_stats} shows that the number of posts for every cause 
varies widely, 
perhaps signifying that 
causes of mental health disorders are not well-distributed in society.

\begin{table}[h]
    \centering
   
    \begin{tabular}{p{0.12\textwidth}p{0.05\textwidth}p{0.06\textwidth}p{0.06\textwidth}p{0.065\textwidth}}
    \hline
    \textbf{Cause} & \textbf{CC} & \textbf{Train\_S} & \textbf{Test\_S}  & \textbf{CAMS}\\
    \hline
    {No reason} & 292 &  332  & 70  & 694   \\
  
   {Bias or abuse} & 122 & 194  & 35  & 351 \\
  
    {Jobs/careers} & 399 & 181 & 48  & 628 \\
    {Medication} & 410 & 170 & 43  & 623 \\
    {Relationship} & 956 & 297 & 91  & 1344 \\
    {Alienation} & 976 &  340 & 92  & 1408 \\
    \hline
    \textbf{Total} & \textbf{3155} & \textbf{1517}   & \textbf{379} & \textbf{5051}\\
    \hline
    \end{tabular}
    \caption{Sample distribution of the CAMS dataset for different causes where \textit{CC} is Crawled Corpus, \textit{Train\_S} is the Training data of SDCNL dataset, \textit{Test\_S} is the Test data of SDCNL dataset, and \textit{CAMS} column contains the total number of samples in the dataset for each cause. }
    \label{tab:cause_stats}
\end{table}

In the crawled corpus, the highest number of samples 
is observed for the 
\textit{Relationship} and \textit{Alienation} causal categories, which is perhaps an indicator that
our society 
is less equipped to deal with 
issues pertaining to \textit{`near \& dear ones'}  and \textit{`loneliness / worthlessness'}, respectively. The number of posts with \textit{`no reason'} is smaller
in the 
crawled
corpus due to the cleaning of the dataset. 
Interestingly, 
there are fewer posts assigned
\textit{`Bias or abuse'}---less than half of each of the two additional categories:
\textit{`Jobs and careers'} and \textit{`Medication'}.

\subsubsection{Social nature of the dataset}
Our expert clinical psychologists have explored the social nature of the dataset, in light of our analysis above. During re-annotation of existing dataset, the prevalence of some causes, e.g., \textit{Alienation} and \textit{Relationship},
point to the importance of the ability to take a societal pulse on a regular basis, especially in these unprecedented times of pandemic-induced distancing and shut-downs.
Other problems, e.g., \textit{Jobs and careers} and \textit{bias and abuse} depend upon good governance. The problem of \textit{medication} depends upon
technological/medical advances and accessible healthcare, or lack thereof.
Additionally, online users often feel depressed but do not address any specific reason behind it, indicating that inferring relevant causes is a challenge if one uses NLP alone.

\subsection{Ethical Considerations}
We emphasize that the sensitive nature of our work necessitates that we use
the publicly available Reddit dataset~\cite{haque2021deep} in a purely observational manner\cite{broer2020technology}. We claim that the given dataset does not disclose the user's personal information or identity. We further acknowledge the trade-off between privacy of data and effectiveness of our work \cite{eskisabel2017ethical}. We ensure that our CAMS corpus is shared selectively and is subject to IRB approval to avoid any misuse. Our dataset is susceptible to the biases and prejudices of annotators who were trained by experts. There will be no ethical issues or legal impact with this causal analysis of mental illness.

\begin{table*}[]
    \centering
    
    \begin{tabular}{cccccccc}
         \hline
         \textbf{Classifier} & \textbf{F1: C0} & \textbf{F1: C1} & \textbf{F1: C2} & \textbf{F1: C3} & \textbf{F1: C4} & \textbf{F1: C5} & \textbf{Accuracy}\\
         \hline
         LR & \textbf{0.63} & \textbf{0.28} & 0.54 & \textbf{0.46} & 0.46 & \textbf{0.53} & \textbf{0.5013}\\
         SVM & 0.54 & 0.23 & \textbf{0.56} & 0.44 & \textbf{0.48} & 0.45 & 0.4670\\
         \hline
         LSTM & 0.54 & \textbf{0.27} & 0.52 & 0.46 & 0.42 & \textbf{0.51} & 0.4595\\
         CNN & 0.56 & \textbf{0.27} & 0.51 & 0.42 & 0.46 & 0.38 & 0.4378\\
         GRU & 0.51 & \textbf{0.27} & 0.54 & \textbf{0.47} & 0.48 & 0.42 & 0.4541\\
         Bi-LSTM & 0.55 & 0.12 & 0.41 & 0.23 & 0.44 & 0.50 & 0.4351\\
         Bi-GRU & \textbf{0.57} & 0.14 & \textbf{0.55} & 0.46 & 0.49 & 0.39 & 0.4568\\
         CNN+GRU & 0.51 & 0.14 & 0.49 & 0.36 & 0.27 & 0.45 & 0.4027\\
         CNN+LSTM & 0.54 & 0.22 & 0.54 & \textbf{0.47} & \textbf{0.54} & 0.47 & \textbf{0.4778}\\
         \hline
    \end{tabular}
    
    \caption{Experimental results with CAMS dataset. F1 is computed for all six causal classes:  `No reason' (C0), `Bias or abuse' (C1), `Jobs and careers' (C2),  `Medication' (C3), `Relationship' (C4), `Alienation' (C5). 
    }
    \label{tab:results}
\end{table*}
\section{Corpus Utility for Machine Learning}
\label{sec:corpus_utility}
We 
include traditional multi-class classifiers trained on CAMS training dataset and evaluate it on the CAMS test data.
We choose the following
Recurrent Neural Network (RNN) architectures: 
Long Short Term Memory (LSTM) model, Convolution Neural Network (CNN), Gated Recurrent Unit (GRU), Bidirectional GRU/LSTM (Bi-GRU/Bi-LSTM) and other hybrid models. In this section, we discuss the experimental setup (5.1)
and analyze the results (5.2).


\subsection{Experimental Setup}
\label{sec:expset}
We use the existing re-annotated SDCNL dataset for experimental results and analysis. We 
clean the dataset, pre-process the posts and then use 
GloVe\footnote{https://nlp.stanford.edu/projects/glove/} word embedding with 100 dimensions trained on Wikipedia for each token. We further set up RNN architectures
with default settings ($lr=0.001, \beta_1=0.9, \beta_2=0.999, \epsilon=1e-08$) for the batch-size of 256. The categorical \textit{Cross Entropy} loss function and the \textit{ADAM} optimizer are used to perform the back-propagation learning on $20$ epochs. 

\mbox{~~~}
The number of samples for three classes of the existing SDCNL dataset (\textit{`Bias or abuse', `Jobs and careers', and `Medication'}) are very few
in comparison to 
the other three classes. We add
$120 + 120 + 120$ samples from the crawled corpus,
to help balance the number of instances across the classes.
As a result, the number of training samples increases from $1517$ to $1877$. We use these training data to build and validate the multi-class classifier. We 
test this classifier on the
$379$ sample test data and analyze the results.
The evaluation metrics used for this task 
are \textit{F1-measure} and \textit{Accuracy}.

\subsection{Results and Discussion}
\label{sec:results}
We use multi-class classifiers to find causal categories and 
obtain the results 
reported in Table~\ref{tab:results}. We test our performance with both machine learning and deep learning approaches. The two top-performing machine learning algorithms are based on \textit{Logistic Regression} and \textit{Support Vector Machine}. Whereas the former outperforms all existing techniques, the latter shows comparative results with deep learning models.
\mbox{~~~}The hybrid model, \textit{CNN-LSTM}, attains the best performance among all deep learning mechanisms with $47.78\%$ accuracy. 
The \textit{CNN+GRU} model performs the worst with $40.27\%$ accuracy on test data. It is interesting to observe that the results are consistent for all the classifiers with few exceptions for classes \textit{Bias or abuse} and \textit{Medication}. We further 
analyze
the best performing deep-learning classifier \textit{(CNN+LSTM)} 
below.

\subsubsection{Error Analysis}
\begin{figure}
    \centering
    \includegraphics[width=0.46\textwidth]{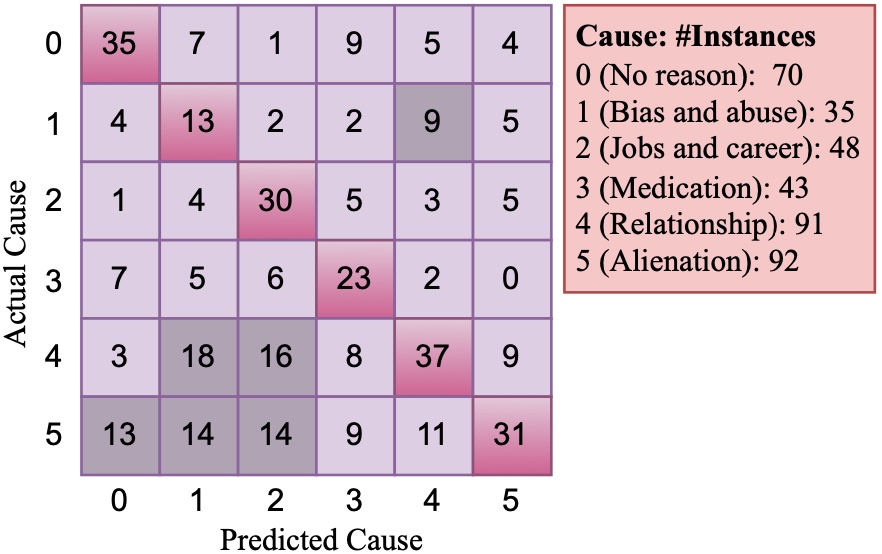}
    \caption{Confusion matrix of CNN+LSTM model on test data}
    \label{fig:con_mat}
\end{figure}

The accuracy 
of multi-class classification is found to be around $40\%$ to $50\%$. 
We undertake a comprehensive error analysis to 
explore
the intricacies of our task.   
\begin{enumerate}
    \item \textit{Cause classification error}: We obtain the confusion matrix for \textit{CNN+LSTM} as shown in Figure~\ref{fig:con_mat}. We highlight the cells with more than $40\%$ incorrect predictions. The predictions for \textit{Alienation} and \textit{Relationship} are incorrect and overlap with \textit{Bias or Abuse} and \textit{Medication}. This is due to complex interactions, as illustrated in the following perceivable overlap between 
    \textit{Bias or Abuse} 
    and
    \textit{Relationship}:
    \begin{quote}
        Example 4: My friends are ignoring me and I am feeling bad about it. I have lost all my friends.
    \end{quote}
    
    Example 4 is associated with \textit{biasing} and \textit{friendship}, 
    in a case where someone feels ostracized by their friends.
    The emphasis on \textit{friends} tips the balance in favor of the class \textit{Relationship}. However, the major challenge is to train the model in such a way that it understands the inferences and then chooses the most emphasized \textit{causal category} using optimization techniques. 
    We view this challenge as an open research direction. 
    \item \textit{Overlapping class}: The \textit{overlapping problem} of classes is observed with ambiguous results for samples, e.g.,
    for \textit{Relationship} and \textit{Alienation}. This class representation problems can be resolved with data augmentation~\cite{ansari2021data} and demarcation of boundaries among classes. In a real-time scenario, demarcation of fixed boundaries 
    is not possible due to subjectivity of the task. We recommend the approximation of a newly built model over handcrafted / automated features accordingly.
    
    \item \textit{Uncertainty}: Related to the overlapping class issue above, all learning-based models obtain low performance for \textit{class 1 (`Bias or abuse')} due to annotators' perceived overlap with classes 4 and 5 (\textit{`Relationship' and `Alienation'}). Future work is needed to mitigate such uncertainty. For example, delineation of discourses within the text would support a more definitive interpretation and reliable annotation.
    
    \item \textit{Semantic Parsing}: In a
    multi-class classification 
    task,
    as the length of posts varies over a wide range, one may choose to 
    summarize 
    every post before applying a classifier. 
    Our
    experiments with YAKE~\cite{campos2020yake} 
    for keyword extraction
    yielded results that are 
    further deteriorated. From this we determine that it
    is important to identify the \textit{causal interpretation} 
    from the full text 
    in order to perform multi-class classification. 
    A future avenue of research involves the exploration of \textit{discourse relations} to identify segments 
    that
    represent
    independent causes
    that underlie mental health disorders.
    
\end{enumerate}

\subsubsection{Implications and Limitations}
The CAMS dataset 
provides a means for exploring the identification of
reasons behind mental
health
disorders of online users. The notion of \textit{causal categorization} 
is defined and used
to proactively identify 
cases where
users are at potential risk of mental depression and suicidal tendencies. The
results of this work may be employed to explore the impact of
\textit{unemployment}, \textit{low grades}, etc. 
Our analysis may also be useful for
the study of online behavior.
A major limitation of the CAMS dataset is that the users may intentionally post their intent of mental disorder on social media, e.g., for deliberately making new friends. In this work, we have assumed that the data has no such biasing.

\section{Conclusion}
\label{sec:conclusion}
This paper introduces
the task of causal analysis to identify the reasons behind mental \textit{depression and suicidal tendencies} (intent). We have introduced CAMS: a dataset of 5051 instances, to categorize the \textit{direct causes} of mental disorders 
through mentions by users in their posts.
We transcend the work of
Level 0 studies, 
moving to the 
next level (Level 1) for causal analysis. Our work is the combined effort of experts in the field of Social Natural Language Processing (Social NLP), including a rehabilitation 
counselor
and clinical psychologists (CPsych). We have further implemented 
machine learning and deep learning models for causal analysis and found that \textit{Logistic Regression} and \textit{CNN+LSTM} gives the best performance, respectively. In the future, we plan to extend the problem of causal analysis 
of mental health detection on social media as a multi-task problem. 
Another major future challenge for this work is the generation of
explanations for multi-class classification, by leveraging
causal analysis within the CAMS framework.
\section{Acknowledgement}
We acknowledge our three student annotators: Simran jeet Kaur, Astha Jain and Ritika Bhardwaj. We also acknowledge Amrith Krishna for his kind support and for proofreading this manuscript. Publication costs are funded by NSERC Discovery Grant (RGPIN-2017-05377), held by Vijay Mago, Department of Computer Science, Lakehead University, Canada.

\section{References}
\label{lr:ref}
\bibliography{LREC22.bbl}
\end{document}